\documentclass[conference,letterpaper]{IEEEtran}

\usepackage[utf8]{inputenc} 
\usepackage{graphicx}
\usepackage[T1]{fontenc}
\usepackage{url}
\usepackage{hyperref}
\usepackage{xcolor}
\usepackage{ifthen}
\usepackage{cite}
\usepackage{amssymb}
\usepackage{amsthm}
\usepackage[cmex10]{amsmath} 

\newcommand{\etal}{\textit{et al.}}

\newcommand{\ERM}{{\sf{ERM}}}

\newcommand{\E}[1]{\mathbb E\{#1\}}
\newcommand{\Esub}[2]{\mathbb E_{#1}\{#2\}}
\newcommand{\gen}{\textup{gen}}

\DeclareMathOperator*{\argmin}{argmin}

\newtheorem{theorem}{\bf{Theorem}}

\newtheorem{corollary}{\bf{Corollary}}

\newtheorem{remark}{\bf{Remark}}

\begin{document}
\title{Information-theoretic analysis for transfer learning} 


\author{%
  \IEEEauthorblockN{Xuetong Wu$^1$, Jonathan H. Manton$^1$, Uwe Aickelin$^2$, Jingge Zhu$^{1}$}
  \IEEEauthorblockA{$^1$Department of Electrical and Electronic Engineering\\
                    $^2$Department of Computing and Information Systems\\
                    University of Melbourne\\ 
                    Parkville, Victoria, Australia\\
                    Email: xuetongw1@student.unimelb.edu, \{jmanton, uwe.aickelin, jingge.zhu\}.unimelb.edu.au}
}


\maketitle

\begin{abstract}
Transfer learning, or domain adaptation, is concerned with  machine learning problems in which training and testing data come from possibly different distributions (denoted as $\mu$ and $\mu'$, respectively).  In this work, we give an information-theoretic analysis on the generalization error and the excess risk of transfer learning algorithms, following  a line of work initiated by Russo and Zhou.  Our results suggest, perhaps as expected, that the Kullback-Leibler (KL) divergence $D(\mu||\mu')$ plays an important role in characterizing the generalization error in the settings of domain adaptation. Specifically, we provide generalization error upper bounds for general transfer learning algorithms, and extend the results to a specific empirical risk minimization (ERM) algorithm where data from both distributions are available in the training phase.  We further apply the method to iterative, noisy gradient descent algorithms, and obtain upper bounds which can be easily calculated, only using parameters from the learning algorithms. A few illustrative examples are provided to demonstrate the usefulness of the results. In particular, our bound is tighter in specific classification problems than the bound derived using Rademacher complexity.
\end{abstract}



\section{Introduction}
Most machine learning methods focus on the setup where the training and testing data are drawn from the same distribution.  Transfer learning, or domain adaptation, is concerned with machine learning problems where training and testing data come from possibly different distributions. This setup is of particular interest in  real-world applications, as in many cases we often have easy access to a substantial amount of data from one distribution, on which our learning algorithm trains, but wish to use the learnt hypothesis for data coming from a different distribution, from which we have limited data for training. 



Generalization error is defined as the difference between the empirical loss and the population loss (defined as (\ref{eq:pop_risk}) and (\ref{eq:erm_risk}) in Section II) for a given hypothesis, and indicates if the hypothesis has been overfitted (or underfitted). Recently, \cite{russo_controlling_2016} proposed an information-theoretic framework for analyzing generalization error of learning algorithms, and showed that the mutual information between the training data and the output hypothesis can be used to upper bound the generalization error. One nice property of this framework is that the mutual information bound explicitly explores the dependence between training data and the output hypothesis, in contrast to the bounds obtained by traditional methods with VC dimension and Rademacher complexity \cite{shalev_shwartz_understanding_2014}. As pointed out by \cite{asadi_chaining_2018}, the information-theoretic upper bound could be substantially tighter than the traditional bounds if we could exploit specific properties of the learning algorithm. While upper bounds on generalization error are classical results in statistical learning theory, only a relatively small number of papers are devoted to this problem for transfer learning algorithms. To mention a few, Ben-David \etal\cite{ben-david_theory_2010} gave VC dimension-style bounds for classification problems. Blitzer \etal\cite{blitzer_learning_2008} and Zhang\cite{zhang2012generalization} studied similar problems and obtained upper bounds in terms of Rademacher complexity.  Specific error bounds for particular transfer learning algorithms and loss metrics are investigated in \cite{dai_boosting_2007} and \cite{zhang_bridging_2019}. Long \etal \cite{long_adaptation_2013} developed a more general framework for transfer learning where the error is bounded with the distribution difference and output hypothesis adaptability.  

Compared with traditional learning problems, the generalization error of transfer learning additionally takes the distribution divergence between the source and target into account and how to evaluate this "domain shift" is non-trivial. We exploit the information-theoretic framework in the transfer learning settings to address this issue following the information-theoretic framework studied by \cite{russo_controlling_2016}, \cite{xu_information-theoretic_2017} and \cite{bu_tightening_2019}. The main contributions are summarized as follows.


\begin{itemize} 
\item[1.] We give an information-theoretic upper bound on the generalization error of transfer learning algorithms where training and testing data come from different distributions and KL-divergence between the source and target distribution captures the effect of domain shift.

\item[2.] We give upper bounds to the excess risk of a specific ERM algorithm where data from both distributions are available to the learning algorithm. Our example shows that our bound is tighter than the existing bounds in specific classification problems which depend on the Rademacher complexity of the hypothesis space, as our bounds are data-algorithm dependent.

\item[3.] We further develop generalization error and excess risk upper bounds for noisy, iterative gradient descent algorithms. The results are useful in the sense that the bounds on the mutual information can be easily calculated only using parameters from the optimization algorithms. 
\end{itemize}

\section{Problem formulation and main results}
We consider an instance space $\mathcal Z$,  a hypothesis space $\mathcal W$ and a non-negative  loss function $\ell: \mathcal W\times\mathcal Z\mapsto \mathbb R^+$. Let $\mu$  and $\mu'$ be two probability distributions defined on $\mathcal Z$, and  assume that $\mu$ is \emph{absolute continuous} with respect to $\mu'$ ($\mu \ll \mu'$). In the sequel, the distribution $\mu$ is referred to as the \textit{source distribution}, and $\mu'$ as the \textit{target distribution}. We are given a set of training data $\{Z_1,\ldots, Z_n\}$. More precisely, for a fixed number $\beta\in[0,1)$, we assume that the samples  $S'=\{Z_1,\ldots, Z_{\beta n}\}$ are drawn IID from the target distribution, and the samples $S=\{Z_{\beta n+1},\ldots, Z_{n}\}$ are drawn IID from the source distribution. 

In the setup of transfer learning, a learning algorithm is  a (randomized) mapping from the training data $S, S'$  to a hypothesis $w\in\mathcal W$, characterized by a conditional distribution $P_{W|SS'}$, with the goal to find a hypothesis $w$ that minimizes the population risk with respect to the \textit{target distribution}
\begin{align}
L_{\mu'}(w):=\Esub{Z\sim\mu'}{\ell(w, Z)}
\label{eq:pop_risk}
\end{align}
where $Z$ is distributed according to $\mu'$.  Notice that  $\beta=0$ corresponds to the important case when we do not have any samples from the target distribution. Obviously, $\beta=1$ takes us back to the classical setup where training data comes from the same distribution as test data, which is not our focus.

\subsection{Empirical risk minimization}\label{subsec:erm}
In this section, we  focus on one particular  \textit{empirical risk minimization} (ERM) algorithm.  For a hypothesis $w\in\mathcal W$, the empirical risk of $w$ on a training sequence $\tilde S:=\{Z_1,\ldots, Z_m\}$ is defined as
\begin{align}
\hat L(w,\tilde S):=\frac{1}{m}\sum_{i=1}^m\ell(w,Z_i).
\label{eq:erm_risk}
\end{align}
Given samples $S$ and $S'$ from both distributions, it is natural to form an empirical risk function as a convex combination of the empirical risk induced by $S$ and $S'$ \cite{ben-david_theory_2010}  defined as
\begin{align*} 
\hat L_{\alpha}(w,S,S'):=\frac{\alpha}{\beta n}\sum_{i=1}^{\beta n}\ell(w,Z_i)+ \frac{1-\alpha}{(1-\beta)n}\sum_{i=\beta n+1}^{n}\ell(w,Z_i)  
\end{align*}
for some weight parameter $\alpha\in[0,1]$ to be determined. We define $W_{\ERM} :=\text{argmin}_w \hat L_{\alpha}(w)$ as the ERM solution,  and also define the optimal hypothesis (with respect to the distribution $\mu'$) as
$w^*=\text{argmin}_{w\in\mathcal W} L_{\mu'}(w)$.

We are interested in two quantities for this  ERM algorithm.  The first one is the  \textit{generalization error} defined as
\begin{align}
\gen(W_{\ERM}, S, S'):= L_{\mu'}(W_{\ERM})-\hat L_{\alpha}(W_{\ERM},S,S')
\label{eq:gen_error}
\end{align}
namely the difference between the minimized empirical risk and the population risk of the ERM solution under the target distribution. We are also interested in the \textit{excess risk} as
\begin{align*}
R_{\text{excess}}(W_{\ERM}) :=L_{\mu'}(W_{\ERM})-L_{\mu'}(w^*)
\end{align*}
which is the difference between the population risk of $W_{\ERM}$ compared to that of the optimal hypothesis. Notice that the excess risk is related to the generalization error via the following upper bound
\begin{align}
L_{\mu'}(W_{\ERM}) & - L_{\mu'}(w^*) \leq  \gen(W_{\ERM}, S, S')+\hat L_{\alpha}(w^*,S,S') \nonumber \\
& -L_{\alpha}(w^*)+(1-\alpha)(L_{\mu}(w^*)-L_{\mu'}(w^*)) \label{eq:excess_expression}
\end{align}
where we have used the fact  $\hat L_{\alpha}(W_{\ERM},S,S')-\hat L_{\alpha}(w^*,S,S')\leq 0$ by the definition of $W_{\ERM}$. For any $w\in\mathcal W$, the quantity $L_{\alpha}(w)$ in the above expression   is defined as
\begin{align*}
L_{\alpha}(w) := (1-\alpha)\Esub{Z\sim\mu}{\ell(w,Z)}+\alpha \Esub{Z\sim\mu'}{\ell(w,Z)}.
\end{align*}

\subsection{Upper bound on generalization errors}

We  view the ERM solution $W_{\ERM}$ as a random variable induced by the random samples $S, S'$ and the (possibly random) ERM algorithm, characterized by a conditional distribution $P_{W|SS'}$. We will first study the expectation of the generalization error
\begin{align}
\Esub{WSS'}{L_{\mu'}(W_{\ERM})-\hat L_{\alpha}(W_{\ERM},S, S')}
\label{eq:generalization_error_expect}
\end{align}
where the expectation is taken with respect to the distribution $P_{WSS'}$ defined as
\begin{align*}
P_{WSS'}(w,z^n) :=P_{W|SS'}(w|z^n)\prod_{i=1}^{\beta n} \mu'(z_i)\prod_{i=\beta n+1}^{n}\mu(z_i).
\end{align*}
Furthermore we use $P_W$ to denote the marginal distribution of $W$ induced by the joint distribution $P_{WSS'}$.

Following the characterization used in \cite{bu_tightening_2019}, the following theorem provides an upper bound on the expectation of the generalization error in terms of the mutual information between individual samples $Z_i$ and the any solution $W$, as well as the KL-divergence between the source and target distributions. As pointed out in \cite{bu_tightening_2019},  using mutual information between the hypothesis and individual samples $I(W;Z_i)$ in general gives a tighter upper bounds than using  $I(W;S)$.
\begin{theorem}[Generalization error of ERM]
Assume that the cumulant generating function of the random variable $\ell(W, Z)-\E{\ell(W,Z)}$  is upper bounded by $\psi(\lambda)$ in the interval $(b_{-},b_{+})$ under the product distribution $P_W\otimes\mu'$ for some $b_{-}<0$ and $b_{+}>0$.  Then for any $\beta>0$, the expectation of the  generalization error  in (\ref{eq:generalization_error_expect}) is upper bounded as
\begin{small}
\begin{align*}
\Esub{WSS'}{\gen(W_\ERM, S, S')}\leq \frac{\alpha}{\beta n}\sum_{i=1}^{\beta n}\psi^{*-1}_{-}(I(W_\ERM;Z_i)) \\
+\frac{(1-\alpha)}{(1-\beta)n}\sum_{i=\beta n+1}^{n}\psi^{*-1}_{-}(I(W_\ERM;Z_i)+D(\mu||\mu'))\\
-\Esub{WSS'}{\gen(W_\ERM, S, S')}\leq \frac{\alpha}{\beta n}\sum_{i=1}^{\beta n}\psi^{*-1}_{+}(I(W_\ERM;Z_i))\\
+\frac{(1-\alpha)}{(1-\beta)n}\sum_{i=\beta n+1}^{n}\psi^{*-1}_{+}(I(W_\ERM;Z_i)+D(\mu||\mu'))
\end{align*}
\end{small}
where we define
\begin{small}
\begin{align*}
\psi^{*-1}_{-}(x) &:=\inf_{\lambda\in[0,-b_{-})}\frac{x+\psi(-\lambda)}{\lambda} \\
\psi^{*-1}_{+}(x) &:=\inf_{\lambda\in[0,b_{+})}\frac{x+\psi(\lambda)}{\lambda}
\end{align*}
\end{small}
\label{thm:exp_gen}
\end{theorem}
All the proofs in this paper can be found in \cite{Proofs_2020}. 
\begin{remark}
In fact, the bound above is not specific to the ERM algorithm, but applicable to {\bf any} hypothesis $W$ generated by a learning algorithm characterized by the conditional distribution $P_{W|S,S'}$ (see proofs in \cite{Proofs_2020} for more details). 
\end{remark}
From a stability point of view \cite{bousquet_stability_2002}, good algorithms (ERM, for example) should ensure that $I(W;Z_i)$ vanishes as $n\rightarrow\infty$. On the other hand, the domain shift is reflected in the KL-divergence $D(\mu||\mu')$, as this term does not vanish when $n$ goes to infinity.

Optimizing $\alpha$ in the above expression is non-trivial as $W_{\ERM}$ inexplicitly involves $\alpha$. However, if we care about the generalization error with respect to the population risk under the target distribution for $n\rightarrow\infty$ (the number of samples  $S'$ from the target distribution also goes to infinity), the intuition says that we should choose $\alpha=1$, i.e. only using $S'$  from the target domain in the training process. On the other hand, if we only have limited data samples, $\alpha$ can be set to be $\beta$ as suggested in \cite{zhang2012generalization,ben-david_theory_2010} that this choice is shown to achieve the faster convergence rate and tighter bound. Overall, we suggest that $\alpha$ should approach 1 with $n$ increasing, say, $\alpha = 1 - O(\frac{1}{n})$.

The result in Theorem \ref{thm:exp_gen} does not cover the case $\beta=0$ (no samples from the target distribution).  However, it is easy to see that in this case we should choose $\alpha=0$ in our ERM algorithm, and a corresponding upper bound is given as in the following corollary under generic hypothesis. 
\begin{corollary}[Generalization error with source only]
Let $\beta=0$ so  that we only have samples $S$ from the source distribution $\mu$. Let $P_{ W|S}$ be the conditional distribution characterizing the learning algorithm which maps samples $S$ to a hypothesis $ W$.(In particular, W is not necessarily the same as $W_{\ERM} = \argmin_{w}\hat{L}(w,S)$). Under the assumptions in Theorem \ref{thm:exp_gen},  the expected generalization error of $ W$ is upper bounded as
\begin{small}
\begin{align*}
\Esub{ WS}{L_{\mu'}( W)-\hat L(W, S)}\leq \frac{1}{n} \sum_{i=1}^{n}\psi^{*-1}_{-}(I( W;Z_i)+D(\mu||\mu'))\\
-\Esub{ WS}{L_{\mu'}( W)-\hat L(W, S)}\leq \frac{1}{n} \sum_{i=1}^{n}\psi^{*-1}_{+}(I( W;Z_i)+D(\mu||\mu'))
\end{align*}
\end{small}
\label{coro:gen_beta0}
\end{corollary}
If the loss function $\ell(W, Z)$ is $r^2$-subgaussian, namely 
\begin{align*}
\log \E{ e^{\lambda(\ell(W,Z)-\E{\ell(W,Z))}}}\leq \frac{r^2\lambda^2}{2}
\end{align*}
for any $\lambda\in\mathbb R$ under the distribution $P_W\otimes \mu'$, the bound in Theorem \ref{thm:exp_gen} can be further simplified with $\psi^{*-1}(y) =\sqrt{2r^2y}$. In particular, if the loss function takes value in $[a,b]$, then $\ell(W,Z)$ is $\frac{(b-a)^2}{4}$-subgaussian. We give the following corollary for the subgaussian loss function.

\begin{corollary}[Generalization error for subgaussian loss functions]
If $\ell(w,Z)$ is $r^2$-subgaussian  under the distribution $P_W\otimes \mu'$, then the expectation of the  generalization error of the ERM solution in (\ref{eq:generalization_error_expect}) is upper bounded as
\begin{small}
\begin{align*}
|\Esub{WSS'}{\gen(W_{\ERM}, S, S')}| \leq \frac{\alpha\sqrt{2r^2}}{\beta n}\sum_{i=1}^{\beta n}\sqrt{I(W_{\ERM};Z_i)} \\
+\frac{(1-\alpha)\sqrt{2r^2}}{(1-\beta)n}\sum_{i=\beta n+1}^{n}\sqrt{ (I(W_{\ERM};Z_i)+D(\mu||\mu'))}
\end{align*}
\end{small}
If $\beta=0$, for any hypothesis $\hat W$ (not necessarily the ERM solution) induced by $S$ and a learning algorithm $P_{\hat W|S}$ , we have the upper bound
\begin{small}
\begin{align}
|\Esub{\hat WS}{L_{\mu'}(\hat W)-\hat L(\hat W, S)}|\leq \frac{\sqrt{2r^2}}{n} \sum_{i=1}^{n}\sqrt{I(\hat W;Z_i)+D(\mu||\mu')}
\label{eq:subgaussian_beta0}
\end{align}
\end{small}
\label{coro:general_bound}
\end{corollary}
The above result follows directly from Corollary \ref{coro:gen_beta0} and  by noticing that we can set $\psi(\lambda)=\frac{r^2\lambda^2}{2}, , b_{-}=-\infty, b_{+}=\infty$ with the assumption that $\ell(W,Z)$ is $r^2$-subgaussian.

\begin{remark}
Using the chain rule of mutual information and the fact that $Z_i$'s are IID, we can relax the upper bound in (\ref{eq:subgaussian_beta0}) as 
\begin{small}
\begin{align*}
\Esub{\hat WS}{L_{\mu'}(\hat W)-\hat L(\hat W, S)}\leq \sqrt{2r^2\left(\frac{I(\hat W;S)}{n} +D(\mu||\mu')\right)}
\end{align*}
\end{small}
which recovers the result in the \cite{xu_information-theoretic_2017} if $\mu=\mu'$. Moreover, we see that the effect of the ``change of domain"  is simply  captured by the KL divergence  between the source and the target distribution. 
\end{remark}
\subsection{Upper bound on  the excess risk  of ERM}
In this section we focus on the case  $\beta>0$ and   give a data-dependent upper bound on the excess risk defined in (\ref{eq:excess_expression}).  To do this, we first define a $L^1$ distance quantity between the two divergent distributions as
\begin{align}
d_{\mathcal W}(\mu, \mu')=\sup_{w\in\mathcal W} |L_{\mu}(w)- L_{\mu'}(w)|.
\label{eq:empirical_distance}
\end{align}
The following theorem gives a bound for the excess risk.
\begin{theorem}[Excess risk of ERM]
Assume that for any $w\in\mathcal W$, the loss function $\ell(w, Z)$ is $r^2$-subgaussian under the distribution $P_{W} \otimes \mu'$. Then for any $\epsilon>0$ and $\delta>0$, there exists an $n_0$ (depending on $\delta$ and $\epsilon$) such that for all $n\geq n_0$, the following inequality holds with probability at least $1-\delta$  (over the randomness of samples and the learning algorithm),
\begin{small}
\begin{align}
L_{\mu'}(W_{\ERM})-L_{\mu'}(w^*) \leq \frac{\alpha\sqrt{2r^2}}{\beta n}\sum_{i=1}^{\beta n}\sqrt{I(W_{\ERM};Z_i)} \nonumber \\
+\frac{(1-\alpha)\sqrt{2r^2}}{(1-\beta)n}\sum_{i=\beta n+1}^{n}\sqrt{ (I(W_{\ERM};Z_i)+D(\mu||\mu'))}\nonumber \\
+ \sqrt{\frac{\alpha^2}{\beta} + \frac{(1-\alpha)^2}{(1-\beta)}}\sqrt{\frac{2r^2\ln{\frac{2}{\delta}}}{n}} +(1-\alpha)d_{\mathcal W}(\mu, \mu')+\epsilon
\label{eq:excess_ERM}
\end{align}
\end{small}
Furthermore in the case when $\beta=0$ (no samples from the distribution $\mu'$),  the inequality becomes
\begin{small}
\begin{align*}
L_{\mu'}(W_{\ERM})- & L_{\mu'}(w^*) \leq \sqrt{\frac{2r^2\log \frac{2}{\delta}}{n}}+|L_{\mu}(w^*)-L_{\mu'}(w^*)| \\
& + \frac{\sqrt{2r^2}}{n}\sum_{i=1}^{n}\sqrt{ (I(W_{\ERM};Z_i)+D(\mu||\mu'))} + \epsilon
\end{align*}
\end{small}
\label{thm:excess}
\end{theorem}
Note that $d_{\mathcal W}(\mu, \mu')$ is normally known as the integral probability metric, which is challenging to evaluate. Sriperumbudur \etal \cite{sriperumbudur2012empirical} investigated the data-dependent estimation to compute the quantity using Kantorovich metric, Dudley
metric and kernel distance, respectively. Ben-David \etal \cite{ben-david_theory_2010} proposed another evaluation method to resolve the issue for classification problem. We point out that the result in Theorem \ref{thm:excess} is not effective for a class of supervised machine learning problems if $\mu$ is not absolutely continuous with respect to $\mu'$. Specifically when the label $Y$ is a deterministic function of the features $X$, the KL divergence is $D(\mu||\mu')=\infty$, leading to a vacuous bound. To develop an appropriate upper bound to handle such scenarios, we follow the methods in \cite{jiao_dependence_2017} to extend the results by using other types of $\phi$-divergence. In particular, we choose $\phi(x) = |x-1|$, which do not impose the absolute continuity restriction.
\begin{corollary} (Generalization error bound of ERM using $\phi$-divergence)
Assume that for any $w\in\mathcal W$, the loss function $\ell(w, Z)$ is $L_{\infty}$-norm bounded by $\sigma$ under the distribution $P_{W} \otimes \mu'$. Then for any $\epsilon>0$ and $\delta>0$, there exists an $n_0$ (depending on $\delta$ and $\epsilon$) such that for all $n\geq n_0$, the following inequality holds with probability at least $1-\delta$  (over the randomness of samples and the learning algorithm) that
\begin{small}
\begin{align}
& L_{\mu'}(W_{\ERM}) -  L_{\mu'}(w^*) \leq \frac{\alpha\|\sigma\|_{\infty}}{\beta n}\sum_{i=1}^{\beta n}I_{\phi}(W_{\ERM};z_i) \nonumber \\
&+\frac{(1-\alpha)\|\sigma\|_{\infty}}{(1-\beta)n}\sum_{i=\beta n+1}^{n} \left( I_{\phi}(W_{\ERM};z_i) + 2TV(\mu  || \mu')\right) + \epsilon \nonumber
\label{eq:excess-fdiv-ERM}
\end{align}
\end{small}
\label{coro:excess-fdiv-ERM}
where $I_{\phi}(W_{\ERM};z_i) = D_{\phi}(P_{W_{\ERM},z_i}||P_{W_{\ERM}} \otimes P_{z_i})$ is the $\phi$-divergence between the distribution $P_{W_{\ERM},z_i}$ and $P_{W_{\ERM}} \otimes P_{z_i}$ with $D_{\phi}(P||Q) = \int|dP - dQ|$ and $TV(\mu  || \mu') = \frac{1}{2}D_{\phi}(\mu || \mu')$ denotes the total variation distance between the distribution $\mu$ and $\mu'$.
\end{corollary}





\subsection{Generalization error bound for noisy gradient descent algorithm}\
The upper bound obtained in previous section cannot be evaluated directly as it depends on the distribution of the data, which is in general assumed unknown in learning problems. Furthermore,
in most cases, $W_{\ERM}$ does not have a closed-form solution, but obtained by using an  optimization algorithm. In this section, we study the class of optimization algorithms that iteratively update its optimization variable based on both source $S$ and target dataset $S^\prime$. The upper bound derived in this section are useful in the sense that the bound can be easily calculated if the relative learning parameters are given. Specifically, the hypothesis $W$ is represented by the optimization variable of the optimization algorithm, and we use $W(t)$ to denote the variable at iteration $t$. In particular, we consider the following noisy gradient descent algorithm 
\begin{equation}
W(t) = W(t-1) - \eta_t\nabla\hat{L}_\alpha(W(t-1),S,S^\prime) + n(t)
\label{eq:iterative}
\end{equation}
where $W(t)$ is initialized to be $W(0) \in \mathcal{W}$ arbitrarily, $\nabla\hat{L}_{\alpha}$ denotes the gradient of $\hat{L}_{\alpha}$ with respect to $W$, and $n(t)$ can be any noises with the mean value of $0$ and variance of $\sigma^2_tI_d \in \mathbb{R}^d$. A typical example is $n(t) \sim \mathcal{N}(0,\sigma^2_tI_d)$. 




\begin{theorem}[Generalization error of noisy gradient descent]
Assume that $W(T)$ is obtained from (\ref{eq:iterative}) at $T$ iteration, and assume that $\ell(w,Z)$ is $r^2\text{-subgaussian}$ over $P_{W} \otimes \mu'$, and the gradient is bounded, e.g., $\left \| \nabla(\ell(w(t),Z))\right\|_2  \leq  K_{ST}$ for any $w(t)$. then 
\begin{small}
\begin{align}
\mathbb{E}_{w S S^{\prime}} & \left\{\operatorname{gen}\left(W(T), S, S^{\prime}\right)\right\} \notag \leq \alpha \sqrt{\frac{2 r^{2}}{\beta n} \hat{I}(S)} \\
 &+ (1-\alpha)\sqrt{2 r^{2}\left(\frac{\hat{I}(S)}{(1-\beta)n}+D\left(\mu \| \mu^{\prime}\right)\right)} 
\label{eq:T_iteration_gen}
\end{align}
\end{small}
where we define
\begin{small}
\begin{equation}
\hat{I}(S) := \frac{d}{2} \sum_{t=1}^{T}\log \left(2 \pi e \frac{\eta_{t}^{2} K_{ST}^{2}+d \sigma_{t}^{2}}{d}\right)- \sum_{t=1}^{T}h(n_t)
\label{eq:SGDMItarget}
\end{equation}
\end{small}
\label{thm:Tbound}
\end{theorem}
In this bound, we observe that if the optimization parameters (such as $\alpha, \beta, n(t), w(0), T, d$) and loss function are fixed, the generalization error bound is easy to calculate by using the parameters given above. Also note that our assumptions do not require that the noise is Gaussian distributed or the loss function $\ell(w;z)$ is convex, this generality provides a possibility to tackle a wider range of optimization problems. However, in many cases $W(T) (\neq W_{\ERM})$ can not be directly applied to bound the excess risk where (\ref{eq:excess_expression}) does not generally hold. One can further provide an excess risk upper bound by utilizing the proposition 3 in \cite{schmidt_convergence_2011} with the assumption of strongly convex loss function, which guarantees the convergence of hypothesis.

\section{Examples}
In this section, we provide two simple examples to illustrate the upper bounds we obtained in previous sections.
\label{sec:example}
\subsection{Estimating the mean of Gaussian}
We consider an example studied in \cite{bu_tightening_2019}. Assume that  $S$ comes from the source distribution $\mu= \mathcal{N} (m, \sigma^2)$ and $S'$ comes form the target distribution $\mu'=\mathcal{N}(m', \sigma^2)$ where $m \neq m'$.  We define the loss function as
\begin{align*}
\ell(w,z)=(w-z)^2.
\end{align*}

For simplicity we  assume here that $\beta=0$. The empirical risk minimization (ERM) solution is obtained by minimizing $\hat L(w,S):=\frac{1}{ n}\sum_{i=1}^{n}(w-Z_i)^2$, where the solution is given by
\begin{small}
\begin{align*}
W_{\ERM}=\frac{1}{n}\sum_{i=1}^nZ_i
\end{align*}
\end{small}
To obtain the upper bound, we first notice that in this case
\begin{align*}
I(W_{\ERM};Z_i)=\frac{1}{2}\log \frac{n}{n-1}
\end{align*}
for all $i$. It is easy to see that the loss function $\ell(W;Z_i)$ is non-central chi-square distribution $\chi^2(1)$ of $1$ degree of freedom with the variance of $\sigma^2_\ell = \frac{n+1}{n}\sigma^2$. 
Furthermore, the cumulant generating function can be bounded as
\begin{small}
\begin{align*}
\log \mathbb{E}{e^{\lambda(\ell(W;Z_i)-\mathbb{E}{\ell(W;Z_i))}}} \leq \sigma^4_\ell\lambda^2 + \frac{2\lambda^2 \sigma^2_{\ell} (m-m')^2}{1 + 2\lambda\sigma^2_{\ell}}, \text{ for } \lambda > 0
\end{align*}
\end{small}
By Corollary \ref{coro:gen_beta0}, the generalization error bound is given as
\begin{equation*}
\mathbb{E} \{ \gen(W_{\ERM})\} \leq \frac{1}{n}\sum_{i=1}^{n}\psi^{*-1}(I(W_{\ERM};Z_i)+D(\mu||\mu'))
\end{equation*}
By the definition of $\psi^{*-1}(x)$, 
\begin{align*}
    \psi^{*-1}(x) \geq (m - m')^2 + \sigma^4_\ell\lambda + \frac{I(W;Z)}{\lambda}
\end{align*}
We set $\lambda = \sqrt{\frac{I(W;Z_i)}{\sigma^4_\ell}}$ and substitute $I(W_{\ERM};Z_i)$ in the generalization error above, we reach
\begin{align*}
\mathbb{E} \{ \gen(W_{\ERM})\} \leq  2\left(\frac{n+1}{n}\right)\sigma^2\sqrt{\frac{1}{2}\log\frac{n}{n-1}} + 2\sigma^2D(\mu \| \mu')
\end{align*}
where $D(\mu||\mu') = \frac{(m - m')^2}{2\sigma^2}$.
In this case, the generalization error of $W_{\ERM}$ can be calculated exactly to be
\begin{align*}
\mathbb{E} \{ \hat L(W_{\ERM}, S)-L_{\mu'}(W_{\ERM}) \}= \frac{2\sigma^2}{n} + 2\sigma^2 D(\mu||\mu')
\end{align*}
The derived bound approaches $2\sigma^2D(\mu||\mu')$ as $n\rightarrow \infty$ with a decay rate $O(1/\sqrt{n})$. The derived bound captures the bound asymptotically well with a lower rate, which is often the results using Rademacher complexity bound\cite{zhang2012generalization}.

\subsection{Logistic regression transfer}
\begin{figure}[h!]
    \centering
    \includegraphics[width=0.25\textwidth]{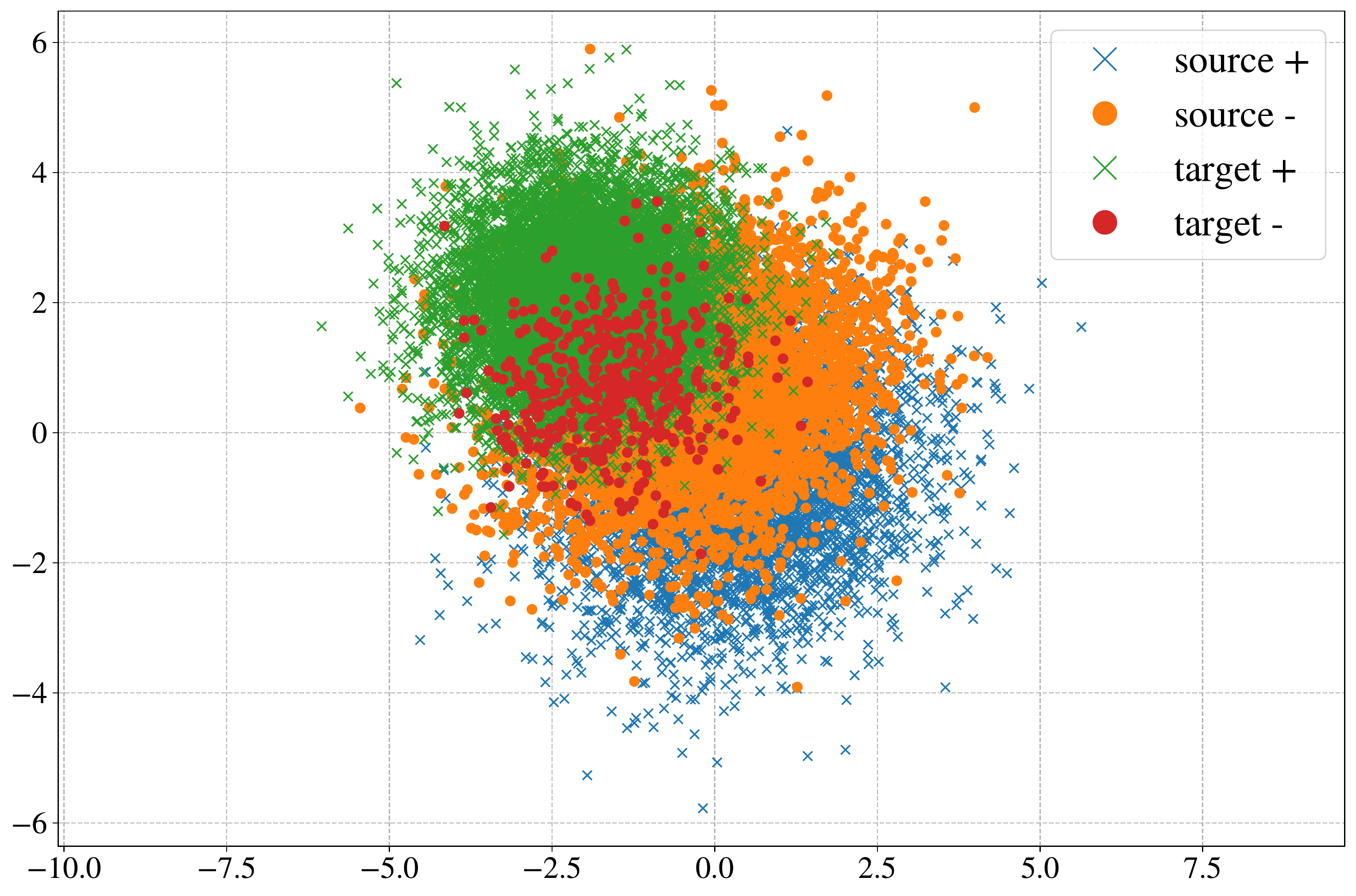}
    \caption{The source data $x_i$ are sampled from the {\bf truncated} Gaussian distribution $\mathcal{N}_{tc} \sim (\mathbf{0},2\mathbf{I})$ while the target data are sampled from the {\bf truncated} Gaussian distribution $\mathcal{N}_{tc} \sim ((-2,2),\mathbf{I})$. The according label $y \in \{0, 1 \}$, is generated from the Bernoulli distribution with probability $p(1) = \frac{1}{1+e^{-w^Tx}}$, where $w_s = (0.5,-1)$ for the source and $w_t = (-0.5,1.5)$ for the target.}
    \label{fig:LogisticRegression}
\end{figure}

In this section, we apply our bound in a typical classification problem. Consider the following logistic regression problem in a 2-dimensional space shown in Figure~\ref{fig:LogisticRegression}. For each $w \in \mathbb{R}^2$ and $z_i = (x_i,y_i) \in \mathbb{R}^{2} \times \{0,1\}$, the loss function is given by
\begin{align*}
    \ell(w,z_i) := -(y_i\log (\sigma(w^Tx_i)) + (1-y_i)\log (1 - \sigma(w^Tx_i)))
\end{align*}
where $\sigma(x) = \frac{1}{1+e^{-x}}$. 

Here we truncate the Gaussian random variables $x_i = \{(x_1,x_2) \big| \|x_1\|_2 < 6, \|x_2\|_2<6 \}$, for $i = 1,\cdots,n$. We also restrict hypothesis space as $\mathcal{W} = \{w: \|w\|_2 < 3\}$ where $W_{\ERM}$ falls in this area with high probability. It can be easily checked that $\mu \ll \mu'$ and the loss function is bounded, hence we can upper bound generalization error using Corollary \ref{coro:general_bound}. To this end, we firstly fix the source samples $n_s = 10000$, while the target samples $n_t$ varies from 100 to 100000 and $\alpha = \beta = \frac{n_t}{n_s + n_t}$ following the guideline from \cite{ben-david_theory_2010, zhang2012generalization}. We give the empirical estimation for $r^2$ within the according hypothesis space such that
\begin{small}
\begin{align*}
      r^2 = \frac{\left(\max_{Z \in S', w \in \mathcal{W}}{\ell(w,Z)} - \min_{Z \in S', w \in \mathcal{W}}{\ell(w,Z)}\right)^2}{4}  
\end{align*}
\end{small}
To evaluate the mutual information $I(W_{\ERM},Z_i)$ efficiently, we follow the work \cite{moddemeijer1989estimation} by repeatedly generating $W_{\ERM}$ and $Z_i$. As $\mu \ll \mu'$ , we decompose $D(\mu(X,Y)\|\mu'(X',Y')) = D(P_X\|P_{X'})+\mathbb{E}_{X \sim P_{X}} \{ D(P_{Y|X=x}\|P_{Y'|X=x})\}$ in terms of the feature distributions and conditional distributions of the labels. The first term $D(P_X\|P_{X'})$ can be calculated using the parameters of Gaussian distributions. The latter term denotes the expected KL-divergence over $P_{X}$ between two Bernoulli distributions, which can be evaluated by generating abundant samples from the source domain. Further we apply Theorem \ref{thm:excess} to upper bound the excess risk, where we give a data-dependent estimation for the term $d_{\mathcal W}(\mu, \mu')$ as
\begin{small}
\begin{align*}
\hat d_{\mathcal W}(\mu, \mu')=\sup_{w\in\mathcal W} |\hat L(w, S)-\hat L(w, S')|.
\end{align*}
\end{small}
To demonstrate the usefulness of our algorithm, we compare the bound in the following theorem using the Rademacher complexity under the same domain adaptation framework. Detailed experiment settings can be found in \cite{Proofs_2020}.


\begin{theorem} (Generalization error of ERM with Rademacher complexity) \cite[Theorem 6.2]{zhang2012generalization}
Assume that for any $w\in\mathcal W$, the loss function $\ell(w, Z)$ is $r^2$-subgaussian under the distribution $P_W \otimes \mu$ or $P_W \otimes \mu'$. Then for any $\delta>0$, the following inequality holds with probability at least $1-\delta$  (over the randomness of samples and the learning algorithm)
\begin{small}
\begin{align*}
\mathbb{E} &\{ \gen(W_{\ERM})\}  \leq (1-\alpha) d_{\mathcal W}(\mu,\mu') +2 \alpha E_{\sigma\otimes \mu}\{ \sup_{w \in \mathcal{W}}\sigma \ell(Z,w) \} \\
& + \frac{2(1-\alpha)}{\beta n} \mathrm{E}_{\sigma}\{\sup_{w \in \mathcal{W}} \sum_{i=1}^{\beta n} \sigma_{i} \ell \left(z_{i}, w \right) \} +3 \alpha \sqrt{\frac{r\ln (4 / \delta)}{\beta n}} \\
& + (1-\alpha) \sqrt{2r^2\ln (\frac{2}{\delta})\left(\frac{\alpha^{2}}{\beta n}+\frac{(1-\alpha)^{2}}{(1-\beta) n}\right)}
\end{align*} 
\end{small}
where $\sigma$ is randomly selected from \{-1,+1\}.
%
%
\label{thm:RC-ER}
\end{theorem}

\begin{figure}[h!]
    \centering
    \includegraphics[width=0.5\textwidth]{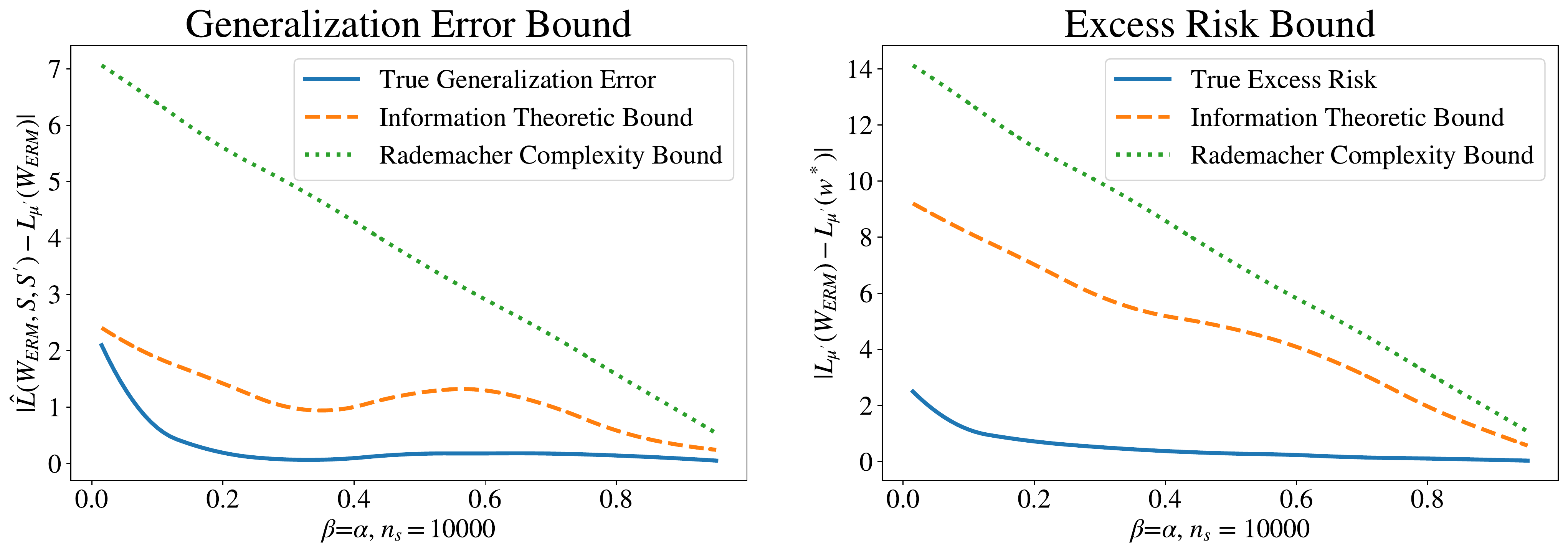}
    \caption{Comparisons for generalization error and excess risk}
    \label{res:LogisticRegression}
\end{figure}

The comparisons of generalization error bound and excess risk bound are shown in figure \ref{res:LogisticRegression}. It is obvious that the true losses are bounded by our developed upper bounds. The result also suggests that our bound is tighter than Rademacher complexity bound in terms of both generalization error and excess risk. This is possibly due to that the generalization error bound with Rademacher complexity is characterized by the domain difference in the whole hypothesis space, while our bound is data-algorithm dependent, which is only concerned with $W_{\ERM}$. As expected, the data-algorithm dependent bound captures the true behaviour of generalization error while Rademacher complexity bound fails to do so. It is noteworthy that both bounds converge as $n$ increases. The result confirms that the bounds captures the dependence of the input data and output hypothesis, as well as the stochasticity of the algorithm.



\bibliographystyle{IEEEtran}
\bibliography{reference}

\IEEEtriggeratref{3}

\end{document}